\documentclass[letterpaper, 10 pt, conference]{ieeeconf}  

\IEEEoverridecommandlockouts                              

\usepackage{amsmath} 
\usepackage{amssymb}  
\usepackage{graphicx}
\usepackage{multirow}
\usepackage{arydshln}  
\usepackage{lettrine} 

\usepackage[hyphens]{url}

\usepackage[hidelinks]{hyperref}   
\usepackage{hyperref}
\usepackage{subcaption}
\usepackage{epstopdf}
\usepackage{graphicx}
\usepackage{amsfonts}
\usepackage{enumerate}
\usepackage{varioref}
\usepackage{longtable}
\usepackage{graphicx}
\usepackage{amssymb}
\usepackage{multirow}
\usepackage{multirow}
\usepackage{caption}

\def\b{\ensuremath\boldsymbol}

\title{\LARGE \bf
Feature Selection and Feature Extraction in Pattern Analysis: \\ A Literature Review
}

\author{Benyamin Ghojogh*, Maria N. Samad*, Sayema Asif Mashhadi*, \\Tania Kapoor*, Wahab Ali*, Fakhri Karray, Mark Crowley \\ \\ \{bghojogh, mnsamad, samashha, t2kapoor, wahabalikhan, karray, mcrowley\}@uwaterloo.ca \\ \\ Department of Electrical and Computer Engineering, University of Waterloo, Waterloo, ON, Canada
\thanks{*The first five authors contributed equally to this work.}
}

\begin{document}

\setlength{\abovedisplayskip}{4pt}
\setlength{\belowdisplayskip}{4pt}

\maketitle
\thispagestyle{empty}
\pagestyle{empty}

\begin{abstract}
Pattern analysis often requires a pre-processing stage for extracting or selecting features in order to help the classification, prediction, or clustering stage discriminate or represent the data in a better way. The reason for this requirement is that the raw data are complex and difficult to process without extracting or selecting appropriate features beforehand. This paper reviews theory and motivation of different common methods of feature selection and extraction and introduces some of their applications. Some numerical implementations are also shown for these methods. Finally, the methods in feature selection and extraction are compared.
\end{abstract}

\begin{keywords}
Pre-processing, feature selection, feature extraction, dimensionality reduction, manifold learning.
\end{keywords}

\section{Introduction}

\lettrine[nindent=0em,lines=2]{P}{attern} recognition has made significant progress recently and is used for various real-world applications, from speech and image recognition to marketing and advertisement. Pattern analysis can be divided broadly into classification, regression (or prediction), and clustering, each of which tries to identify patterns in data in a specific way. The module which finds the pattern in data is named ``model'' in this paper. Feeding the raw data to the model, however, might not result in a satisfactory performance because the model faces a hard responsibility to find the useful information in the data. Therefore, another module is required in pattern analysis as a pre-processing stage to extract the useful information concealed in the input data. This useful information can be in terms of either better representation of data or better discrimination of classes in order to help the model perform better. There are two main categories of methods for this goal, i.e., feature selection and feature extraction. 

To formally introduce feature selection and extraction, first, some notations should be defined. Suppose that there exist $n$ data samples (points), denoted by $\{\b{x}_i\}_{i=1}^n$, from which we want to select or extract features.
Each of these data samples, $\b{x}_i$, is a $d$-dimensional column vector (i.e., $\b{x}_i \in \mathbb{R}^d$).
The data samples $\{\b{x}_i\}_{i=1}^n$ can be represented by a matrix $\b{X} := [\b{x}_1, \dots, \b{x}_n] = [\b{x}^1, \dots, \b{x}^d]^\top \in \mathbb{R}^{d \times n}$. 
In supervised cases, the target labels are denoted by $\b{t} := [t_1, \dots, t_n]^\top \in \mathbb{R}^{n}$.
Throughout this paper, $x$, $\b{x}$, $\b{X}$, $\mathtt{X}$, and $\mathcal{S}$ denote a scalar, column vector, matrix, random variable, and set, respectively. 
The subscript ($\b{x}_i$) and superscript ($\b{x}^j$) on the vector represent the $i$-th sample and the $j$-th feature, respectively. Therefore, $x_i^j$ denotes the $j$-th feature of the $i$-th sample.
Moreover, in this paper, $\b{1}$ is a vector with entries of one, $\b{1} = [1, 1, \dots, 1]^\top$, and $\b{I}$ is the identity matrix. The $\ell_p$-norm is also denoted by $||.||_p$.

Both feature selection and extraction reduce the dimensionality of data \cite{khalid2014survey}.
The goal of both feature selection and extraction is mapping $\b{x} \mapsto \b{y}$ where $\b{x} \in \mathbb{R}^{d}$, $\b{y} \in \mathbb{R}^{p}$, and $p \leq d$. In other words, the goal is to have a better representation of data with dimension $p$, i.e., $\b{Y} := [\b{y}_1, \dots, \b{y}_n] = [\b{y}^1, \dots, \b{y}^p]^\top \in \mathbb{R}^{p \times n}$. 
In feature selection, $p \leq d$ and $\{\b{y}^j\}_{j=1}^p \subseteq \{\b{x}^j\}_{j=1}^d$ meaning that the set of selected features is an inclusive subset of the original features. 
However, feature extraction tries to extract a completely new set of features (dimensions) from the pattern of data rather than selecting features out of the existing attributes \cite{sorzano2014survey}. In other words, in feature extraction, the feature set of $\{\b{y}^j\}_{j=1}^p$ is not a subset of features of $\{\b{x}^j\}_{j=1}^d$ but is a different space. In feature extraction, often $p \ll d$ holds.

\section{Feature Selection}

Feature selection \cite{chandrashekar2014survey,miao2016survey,cai2018feature} is a method of feature reduction which maps $\b{x} \in \mathbb{R}^{d} \mapsto \b{y} \in \mathbb{R}^{p}$ where $p < d$. The reduction criterion usually either improves or maintains the accuracy or simplifies the model complexity. When there are $d$ number of features, the total number of possible subsets are ${2}^{d}$. It is infeasible to enumerate the exponential number of subsets if $d$ is large. Therefore, we need to come up with some method that works in a reasonable time. The evaluation of the subsets is based on some criterion, which can be categorized as filter or wrapper methods \cite{chandrashekar2014survey,web_sudeshna_youtube}, explained in the following.

\subsection{Filter Methods}

Consider a ranking mechanism used to grade the features (variables) and the features are then removed by setting a threshold \cite{Guyon2003introduction}. These ranking methods are categorized as filter methods because they filter the features before feeding to a learning model. 
Filter methods are based on two concepts ``relevance'' and ``redundancy'', where the former is dependence (correlation) of feature with target and the latter addresses whether the features share redundant information. 
In the following, the common filter methods are introduced.


\subsubsection{Correlation Criteria}

Correlation Criteria (CC), also known as Dependence Measure (DM), is based on the relevance (predictive power) of each feature. The predictive power is computed by finding the correlation between the independent feature $\b{x}^j$ and the target (label) vector $\b{t}$. The feature with the highest correlation value will have the highest predictive power and hence will be most useful. The features are then ranked according to some correlation-based heuristic evaluation function \cite{hall_thesis}. One of the widely used criteria for this type of measure is Pearson Correlation Coefficient (PCC) \cite{Guyon2003introduction,battiti1994} defined as: 
\begin{equation}
\rho_{\b{x}^j, \b{t}} :=\frac{\mathrm{cov}(\b{x}^j,\b{t})}{\sqrt{\mathrm{var}(\b{x}^j)\,\mathrm{var}(\b{t})}},
\end{equation}
where $\text{cov}(.,.)$ and $\text{var}(.)$ denote the covariance and variance, respectively.

\subsubsection{Mutual Information}

Mutual Information (MI), also known as Information Gain (IG), is the measure of dependence or shared information between two random variables \cite{Thomas2006}. It is also described as Information Theoretic Ranking Criteria (ITRC) \cite{Guyon2003introduction,battiti1994,john1997wrapper}. 
The MI can be described using the concept given by Shannon's definition of entropy:
\begin{equation}\label{equation_entropy}
H(\mathtt{X}) := - \sum_{\mathtt{x}} p(\mathtt{x}) \log\big(p(\mathtt{x})\big),
\end{equation}
which gives the uncertainty in the random variable $\mathtt{X}$. 
In feature selection, we need to maximize the mutual information (i.e., relevance) between the feature and the target variable. The mutual information (MI), which is the relative entropy between the joint distribution and product distribution, is:
\begin{equation}\label{equation_mutual_information}
\text{MI}(\mathtt{X};\mathtt{T}) := \sum_{\mathtt{X}} \sum_{\mathtt{T}} p(\b{x}^j,\b{t}) \log\frac{p(\b{x}^j,\b{t}) }{p(\b{x}^j)p(\b{t})},
\end{equation}
where $p(\b{x}^j,\b{t})$ is the joint probability density function of feature $\b{x}^j$ and target $\b{t}$. The $p(\b{x}^j)$ and $p(\b{t})$ are the marginal density functions.
The MI is zero or greater than zero if $\mathtt{X}$ and $\mathtt{Y}$ are independent or dependent, respectively.
For maximizing this MI, a greedy step-wise selection algorithm is adopted. In other words, there is a subset of features, denoted by matrix $\mathcal{S}$, initialized with one feature and features are added to this subset one by one. Suppose $Y_{\mathcal{S}}$ denotes the matrix of data having features whose indices exist in $\mathcal{S}$. The index of selected feature is determined as \cite{web_mutual_information}:
\begin{equation}
j := \arg \max_{j\not\in \mathcal{S}} \text{ MI}\big(Y_{\mathcal{S}} \cup \b{x}^j; \b{t} \big).
\end{equation}
The above equation is also used for selecting the initial feature.
Note that it is assumed the selected features are independent.
The stop criterion for adding the new features is when there is highest increase in the MI at the previous step.
It is noteworthy that if a variable is redundant or not informative in isolation, this does  not mean that it cannot be useful when combined with another variable \cite{Pablo2014mutual}. 
Moreover, this approach can reduce the dimensionality of features without having any significant negative impact on the performance \cite{nicolosi2008feature}.




\subsubsection{$\chi^2$ Statistics}

The $\chi^2$ Statistics method measures the dependence (relevance) of feature occurrence on the target value \cite{nicolosi2008feature} and is based on the $\chi^2$ probability distribution \cite{george2004metric}, defined as $\mathtt{Q} = \sum_{i=1}^{k} \mathtt{Z}_i^2$, where $\mathtt{Z}_i$'s are independent random variables with standard normal distribution and $k$ is the degree of freedom.
This method is only applicable to cases where the target and features can take only discrete finite values.
Suppose $n_{p,q}$ denotes the number of samples which have the $p$-th value of the $j$-th feature and the $q$-th value of the target $t$. 
Note that if the $j$-th feature and the target can have $\mathtt{p}$ and $\mathtt{q}$ possible values, respectively, then $p = \{1, ..., \mathtt{p}\}$ and $q = \{1, ..., \mathtt{q}\}$.
A contingency table is formed for each feature using the $n_{p,q}$ for different values of target and the feature.
The $\chi^2$ measure for the $j$-th feature is obtained by:
\begin{equation}
\chi^2(\b{x}^j, \b{t}) := \sum_{p=1}^{\mathtt{p}} \sum_{q=1}^{\mathtt{q}} \frac{(n_{p,q} - \mathbb{E}_{p,q})^2}{\mathbb{E}_{p,q}},
\end{equation}
where $\mathbb{E}_{p,q}$ is the expected value for $n_{p,q}$ and is obtained as:
\begin{equation}\label{equation_chi_squared_ecpectedValue}
\mathbb{E}_{p,q} = n \times \text{Pr}(p) \times \text{Pr}(q) = n \times \frac{\sum_{q=1}^{\mathtt{q}} n_{p,q}}{n} \times \frac{\sum_{p=1}^{\mathtt{p}} n_{p,q}}{n}. 
\end{equation}
The $\chi^2$ measure is calculated for all the features.
The large $\chi^2$ shows the significant dependence of the feature to the target; therefore, if it is less than a pre-defined threshold, the feature can be discarded. 
Note that $\chi^2$ Statistics face a problem when some of the values of features have very low frequency. The reason for this problem is that the expected value in equation (\ref{equation_chi_squared_ecpectedValue}) becomes inaccurate at low frequencies.

\subsubsection{Markov Blanket}

Feature selection based on Markov Blanket (MB) \cite{shunkai2010markov} is a category of methods based on relevance. MB considers every feature $\b{x}^j$ and the target $\b{t}$ as a node in a faithful Bayesian network. The MB of a node is defined as the parents, children, and spouses of that node; therefore, in a graphical model perspective, the nodes in MB of a node suffice for estimating that node. In MB methods, the MB of a target node is found, the features in that MB are selected, and the rest are removed. Different methods have been proposed for finding the MB of target, such as Incremental Association MB (IAMB) \cite{tsamardinos2003algorithms}, Grow-Shrink (GS) \cite{margaritis2000bayesian}, Koller-Sahami (KS) \cite{koller1996toward}, and Max-Min MB (MMMB) \cite{tsamardinos2003time}. The MMMB as one of the best methods in MB is introduced here. 
In MMMB, first, the parents and children of target node are found. For that, those features are selected (denoted by $\mathcal{S}$) that have the maximum dependence with the target, given the subset of $\mathcal{S}$ with the minimum dependence with the target. This set of features includes some false positive nodes which are not parent or child of the target. Thus, the false positive features are filtered out if the selected features and the target are independent given any subset of selected features. Note that the $G^2$ test, Fisher correlation, Spearman correlation, and Pearson correlation can be used for the conditional independence test.
Next, spouses of target are found to form the MB with the $S$ (previously found parents and children). For this, the same procedure is done for the nodes in set $S$ to find the spouses, grandchildren, grandparents, and siblings of the target node. Everything except the spouses should be filtered out; thus, if a feature is dependent on the target given any subset of nodes having their common child, it is retained and the rest are removed. The parents, children, and spouses of the target found form the MB.

\subsubsection{Consistency-based Filter}\label{section_consistency_based_filter}


Consistency-Based Filters (CBF) use a consistency measure which is based on both relevance and redundancy and is a selection criterion that aims to retain the discrimination power of the data defined by original features \cite{dash2003consistency}. 
The InConsistency Rate (ICR) of all features is calculated via the following steps. 
First, inconsistent patterns are found; these are defined as patterns that are identical but are assigned to two or more different class labels, e.g., samples having patterns $(0,1)$ and $(0,1)$ in two features and with targets $0$ and $1$, respectively. 
Second, the inconsistency count for a pattern of a feature subset is calculated by taking the total number of times the inconsistent pattern occurs in the entire dataset minus the largest number of times it occurs with a certain target (label) among all targets. For example, assume $c_1$, $c_2$, and $c_3$ are the number of samples having targets $1$, $2$, and $3$, respectively. If $c_3$ is the largest, the inconsistency count will be $c_1+c_2$. 
Third, ICR of a feature subset is the sum of inconsistency counts over all patterns (in a feature subset, there can be multiple inconsistent patterns) divided by the total number of samples $n$. 
The feature subset $\mathcal{S}$ with $\text{ICR}(\mathcal{S}) \leq \delta$ is considered to be consistent, where $\delta$ is a pre-defined threshold. This threshold is included to tolerate noisy data. 
For selecting a feature subset $\mathcal{S}$ in CBF methods, FocusM and Automatic-Branch-and-Bound (ABB) are exhaustive search methods that can be used. This yields the smallest subset of consistent features. Las Vegas Filter (LVF), SetCover, and Quick-Branch-and-Bound (QBB) are faster search methods and can be used for large datasets when exhaustive search is not feasible.


\subsubsection{Fast Correlation-based Filter}


Fast Correlation-Based Filter (FCBF) \cite{yu2003feature} uses an entropy-based measure called Symmetrical Uncertainty (SU) to find correlation, for both relevance and redundancy, as follows:
\begin{equation}\label{equation_Symmetrical_Uncertainty}
\text{SU}(\mathtt{X},\mathtt{Y}) := 2 \big(\frac{\text{IG}(\mathtt{X};\mathtt{Y})}{H(\mathtt{X}) + H(\mathtt{Y})}\big),
\end{equation}
where $\text{IG}(\mathtt{X};\mathtt{Y}) = H(\mathtt{X}) - H(\mathtt{X} | \mathtt{Y})$ is the information gain and $H(\mathtt{X})$ is entropy defined by equation (\ref{equation_entropy}).
The value $\text{SU}(\b{x}^i,\b{t})$ between a feature $\b{x}^i$ and target $\b{t}$ is calculated for each feature. 
A threshold value for SU is used to determine whether a feature is relevant or not. A subset of features $\mathcal{S}$ is decided by this threshold. To find out if a relevant feature $\b{x}^i$ is redundant or not, $\text{SU}(\b{x}^i,\b{x}^j)$ value between two features $\b{x}^i$ and $\b{x}^j$ is calculated. For a feature $\b{x}^i$, all its redundant features are collected together (denoted by $S_{P_i}$) and divided in two subsets, $S_{P_i}^+$ and $S_{P_i}^-$, where $S_{P_i}^+ = \{\b{x}^j | \b{x}^j \in S_{P_i}, \text{SU}(\b{x}^j, \b{t}) > \text{SU}(\b{x}^i, \b{t})\}$ and $S_{P_i}^- = \{\b{x}^j | \b{x}^j \in S_{P_i}, \text{SU}(\b{x}^j, \b{t}) \leq \text{SU}(\b{x}^i, \b{t})\}$. All features in $S_{P_i}^+$ are processed before making a decision on $\b{x}^i$. If a feature is predominant (i.e., its $S_{P_i}^+$ is empty) then all features in $S_{P_i}^-$ are removed and $\b{x}^i$ is retained. The feature with the largest $\text{SU}(\b{x}^i, \b{t})$ value is a predominant feature and used as a starting point to eliminate other features. 
FCBF uses the novel idea of predominant correlation to make feature selection faster and more efficient so that it can be easily used on high dimensional data. This method greatly reduces dimensionality and increases classification accuracy.

\subsubsection{Interact}

In feature selection, many algorithms apply correlation metrics to find which feature correlates most to the target. These algorithms single out features and do not consider the combined effect of two or more features with the target. In other words, some features might not have individual effect but alongside other features they give high correlation to the target and increase classification performance. Interact \cite{zhao2007searching} is a fast filter algorithm that searches for interacting features. 
First, Symmetrical Uncertainty (SU), defined by equation (\ref{equation_Symmetrical_Uncertainty}), is used to evaluate the correlation of individual features with the target. This heuristic is used to rank features in descending order such that the most important feature is positioned at the beginning of the list $\mathcal{S}$. 
Interact also uses Consistency Contribution (cc) or c-contribution metric which is defined as:
\begin{equation}
\text{cc}(\b{x}^i, \b{X}) := \text{ICR}\big( \b{X} \backslash \b{x}^i \big) - \text{ICR}\big( \b{X} \big),
\end{equation}
where $\b{x}^i$ is the feature for which cc is being calculated, and ICR stands for inconsistency rate defined in Section \ref{section_consistency_based_filter}. The ``$\backslash$'' is the set-theoretic difference operator, i.e., $\b{X} \backslash \b{x}^i$ means $\b{X}$ excluding the feature $\b{x}^i$.
The cc is calculated for each feature from the end of the list $\mathcal{S}$ and if the cc of a feature is less than a pre-defined threshold, that feature is eliminated. 
This backward elimination makes sure that the cc is first calculated for the features having small correlation with the target.
The appropriate pre-defined threshold is assigned based on cross validation. The Interact method has the advantage of being fast. 

\subsubsection{Minimal-Redundancy-Maximal-Relevance}

The Minimal Redundancy Maximal Relevance (mRMR) \cite{ding2005minimum,peng2005feature} is based on maximizing the relevance and minimizing redundancy of features. 
The relevance of features means that each of the selected features should have the largest relevance (correlation or mutual information \cite{shirzad2015feature}) with the target $\b{t}$ for having better discrimination \cite{john1997wrapper}. This dependence or relevance is formulated as the average mutual information between the selected features (set $\mathcal{S}$) \cite{peng2005feature}:
\begin{equation}
d := \frac{1}{|\mathcal{S}|} \sum_{\b{x}^i \in \mathcal{S}} \text{MI} (\b{x}^i; \b{t}),
\end{equation}
where MI is the mutual information defined by equation (\ref{equation_mutual_information}). 
The redundancy of features, on the other hand, should be minimized because having at least one of the redundant features suffices for a good performance. The redundancy of features $\b{x}^i$ and $\b{x}^j$ is formulated as \cite{peng2005feature}:
\begin{equation}
r := \frac{1}{|\mathcal{S}|^2} \sum_{\b{x}^i, \b{x}^j \in \mathcal{S}} \text{MI} (\b{x}^i; \b{x}^j).
\end{equation}
By defining $\phi = d - r$, the goal of mRMR is to maximize $\phi$ \cite{peng2005feature}. Therefore, the features which maximize $\phi$ are added to the set $\mathcal{S}$ incrementally and one by one \cite{peng2005feature}.

\subsection{Wrapper Methods}

As seen previously, filter methods select the optimal features to be passed to the learning model, i.e., classifier, regression, etc.
Wrapper methods, on the other hand, integrate the model within the feature subset search. In this way, different subsets of features are found or generated and evaluated through the model. The fitness of a feature subset is evaluated by training and testing it on the model. Thus in this sense, the algorithm for the search of the best suboptimal subset of the feature set is essentially ``wrapped'' around the model. The search for the best subset of the feature set, however, is an NP-hard problem. Therefore, heuristic search methods are used to guide the search. These search methods can be divided in two categories: Sequential and Metaheurisitc algorithms.

\subsubsection{Sequential Selection Methods}

Sequential feature selection algorithms access the features from the given feature space in a sequential manner. These algorithms are called sequential due to the iterative nature of the algorithms. 
There are two main categories of sequential selection methods: Sequential Forward Selection (SFS) and Sequential Backward Selection (SBS) \cite{aha1996comparative}. Although both the methods switch between including and excluding features, they are based on two different algorithms according to the dominant direction of the search.  
In SFS, the set of selected features, denoted by $\mathcal{S}$, is initially empty. The features are added to $\mathcal{S}$ one by one if they improve the model performance the best. This procedure is repeated until the required number of features are selected. In SBS, on the other hand, the set $\mathcal{S}$ is initialized by the entire features and features are removed sequentially based on the performance \cite{chandrashekar2014survey}. The SFS and SBS ignore the dependence of features, i.e., some specific features might have better performance than the case where those features are used alongside some other features. Therefore, Sequential Floating Forward Selection (SFFS) and Sequential Floating Backward Selection (SFBS) \cite{pudil1994floating} were proposed. In SFFS, after adding a feature as in SFS, every feature is tested for being excluded if the performance improves. Similar approach is performed in SFBS but with opposite direction.

\subsubsection{Metaheuristic Methods}

The metaheuristic algorithms are also referred to as evolutionary algorithms. These methods have low implementation complexity and can adapt to a variety of problems. They are also less prone to get stuck in a local optima as compared to sequential methods, where the objective function is the performance of the model.
Many metaheuristic methods have been used for feature selection such as the binary dragonfly algorithm used in \cite{dragonfly2017}. Another recent technique called whale optimization algorithm was used for feature selection and explored against other common metaheuristic methods in \cite{whale2017}. As examples of metaheuristic approaches for feature selection, the two broad metaheuristic methods, Particle Swarn Optimization (PSO) and Genetic Algorithms (GA), are explained here. PSO is inspired by the social behavior observed in some animals like the flocking of birds or formation of schools of fish and GA is inspired by natural selection and has been used widely in search and optimization problems. 

The original version of PSO \cite{canonicalpso1995} was developed for continuous optimization problems whose potential solutions are represented by particles. However, many problems are combinatorial optimization problems usually with binary decisions. Geometric PSO (GPSO) uses a geometric framework described in \cite{moraglio2007geometric} to provide a version of PSO that can be generalized to any solution representation. 
GPSO considers the current position of particle, the global best position, and the local best position of the particle as the three parents of particle and creates the offspring (similar to GA approach) by applying a three-Parent Mask-Based Crossover (3PMBCX) operator on them. The 3PMBCX simply takes each element of crossover with specific probabilities from the three parents.
The GPSO is used in \cite{talbi2008comparison} for the sake of feature selection. 
Binary PSO \cite{kennedy1997discrete} can also be used for feature selection where binary particles represent selection of features.

In GA, the potential solutions are represented by chromosomes \cite{geneticalgo2015}. For feature selection, the genes in the chromosome correspond to features and can take values $1$ or $0$ for selection or not selection of feature, respectively. The generations of chromosomes improve by crossovers and mutations until convergence.  
The GA is used in different works such as \cite{kazemiga2014,hotelroomga2015} for selecting features.
A modified version of GA, named CHCGA \cite{chcga1991}, is used in \cite{chcgause} for feature selection. The CHCGA maintains diversity and avoids stagnation of the population by using a Half Uniform Crossover (HUX) operator which crosses over half of the non-matching genes at random.
One of the problems of GA is poor initialization of chromosomes. A modified version of GA proposed in \cite{jiang2017modified} for feature selection reduces the risk of poor initialization by introducing some excellent chromosomes as well as random chromosomes for initialization. In that work, the estimated regression coefficients $\hat{\beta}_1,\dots,\hat{\beta}_d$ and their estimated standard deviations $\sigma_{\hat{\beta}_1},\dots,\sigma_{\hat{\beta}_d}$ are calculated using regression, then Student's $t$-test is applied for every coefficient $ t_j \leftarrow (\hat{\beta}_j-0)/\sigma_{\hat{\beta}_j}$. The binary excellent chromosome $\b{s} = [s_1, \dots, s_d]^\top$ is generated where $s_j = 1$ with probability $\mid t_j \mid / \sum_{i=1}^{d} \mid t_i \mid$.

\section{Feature Extraction}

As features of data are not necessarily uncorrelated (matrix $\b{X}$ is not full rank), they share some information and thus there usually exists dummy information in the data pattern. Hence, in mapping $\b{x} \in \mathbb{R}^{d} \mapsto \b{y} \in \mathbb{R}^{p}$ usually $p \ll d$ or at least $p < d$ holds, where $p$ is named the intrinsic dimension of data \cite{carreira1997review}. 
This is also referred to as the manifold hypothesis \cite{fefferman2016testing} stating that the data points exist on a lower dimensional sub-manifold or subspace.
This is the reason that feature extraction and dimensionality reduction are often used interchangeably in the literature. 
The main goal of dimensionality reduction is usually either better representation/discrimination of data \cite{burges2010dimension} or easier visualization of data \cite{engel2012survey,liu2017visualizing}. 
It is noteworthy that the $\mathbb{R}^p$ space is referred to as feature space (i.e., \textit{feature extraction}), embedded space (i.e., \textit{embedding}), encoded space (i.e., \textit{encoding}), subspace (i.e., \textit{subspace learning}), lower-dimensional space (i.e., \textit{dimensionality reduction}) \cite{baraniuk2010low}, sub-manifold (i.e., \textit{manifold learning}) \cite{cayton2005algorithms}, or representation space (i.e., \textit{representation learning}) \cite{bengio2013representation} in the literature. 

The dimensionality reduction methods can be divided into two main categories, i.e., supervised and unsupervised methods. Supervised methods take into account the labels and classes of data samples while the unsupervised methods are based on the variation and pattern of data.
Another categorization of feature extraction is dividing methods into linear and non-linear. The former assumes that the data falls on a linear subspace or classes of data can be distinguished linearly, while the latter supposes that the pattern of data is more complex and exists on a non-linear sub-manifold. 
In the following, we explore the different methods of feature extraction. For the sake of brevity, we mention the basics and forgo some very detailed methods and improvements of these methods such as using Nystr{\"o}m Method for incorporating out-of-sample data \cite{strange2014open}. 

\subsection{Unsupervised Feature Extraction}

Unsupervised methods in feature extraction do not use the labels/classes of data for feature extraction \cite{robert2014machine,friedman2001elements}. In other words, they do not respect the discrimination of classes but they mostly concentrate on the variation and distribution of data. 

\subsubsection{Principal Component Analysis}

Principal Component Analysis (PCA) \cite{wold1987principal,abdi2010principal} was first proposed by \cite{pearson1901liii}. As a linear unsupervised method, it tries to find the directions which represent the variation of data the best. 
The original coordinates do not necessarily represent the direction of variation. The aim of PCA is to find orthogonal directions which represent the data with the least error \cite{ghodsi2006dimensionality}. Therefore, PCA can be also considered as rotating the coordinate system \cite{web_Ali_Ghodsi_youtube}. 

The projection of data onto direction $\b{u}$ is $\b{u}^\top \b{X}$. Taking $\b{S}$ to be the $d \times d$ covariance matrix of data, the variance of this projection is $\b{u}^\top \b{S} \b{u}$. PCA tries to maximize this variance to find the most variant orthonormal directions of data.
Solving this optimization problem \cite{ghodsi2006dimensionality} yields to $\b{S} \b{u} = \lambda \b{u}$ which is the eigenvalue problem \cite{ghojogh2019eigenvalue} for the covariance matrix $\b{S}$. Therefore, the desired directions (columns of matrix $\b{U}$) are the eigenvectors of the covariance matrix of data. The eigenvectors, sorted by their eigenvalues in descending order, represent the largest to smallest variations of data and are named principal directions or axes. The features (rows) of the projected data $\b{U}^\top \b{X}$ are named principal components.
Usually, the components with smallest eigenvalues are cut off to reduce the data. There are different methods for estimating the best number of components to keep (denoted by $p$), such as using Bayesian model selection \cite{minka2001automatic}, scree plot \cite{cattell1966scree}, and comparing the ratio $\lambda_i / \sum_{j=1}^d \lambda_j$ with a threshold \cite{abdi2010principal} where $\lambda_i$ denotes the eigenvalue related to the $i$-th principal component.

In this paper, out-of-sample data refers to the data sample which does not exist in the training set of samples from wich the subspace was created. 
Assume $\b{U} = [\b{u}_1, \dots, \b{u}_d] \in \mathbb{R}^{p \times d}$. 
The projection of training data $\b{X}$ and the out-of-sample data $\b{x}$ are $\b{Y} = \b{U}^\top \b{X} \in \mathbb{R}^{p \times n}$ and $\b{y} = \b{u}^\top \b{x}$, respectively.
Note that the projected data can also be reconstructed back but it will have distortion error because the samples were projected onto a subspace. The reconstruction of training and out-of-sample data are $\widehat{\b{X}} = \b{U} \b{Y}$ and $\widehat{\b{x}} = \b{U} \b{y}$, respectively.

\subsubsection{Dual Principal Component Analysis}

The explained PCA was based on eigenvalue decomposition; however, it can be done based on Singular-Value Decomposition (SVD) more easily \cite{ghodsi2006dimensionality}. Considering $\b{X} = \b{U} \b{\Sigma} \b{V}^\top$, the $\b{U}$ is exactly the same as before and contains the eigenvectors (principal directions) of $\b{X}\b{X}^\top$ (the covariance matrix). The matrix $\b{V}$ contains the eigenvectors of $\b{X}^\top \b{X}$.
In cases where $d \gg n$, such as in images, calculating eigenvectors of $\b{X}\b{X}^\top$ with size $d \times d$ is less efficient than $\b{X}^\top \b{X}$ with size $n \times n$. Hence, in these cases, dual PCA is used rather than the ordinary (or direct) PCA \cite{ghodsi2006dimensionality}. 
In dual PCA, projection and reconstruction of training data ($\b{X}$) and out-of-sample data ($\b{x}$) are formulated as:
\begin{align}
& \text{Projection for } \b{X}: && \b{Y} = \b{\Sigma} \b{V}^\top, \label{equation_dualPCA_projection} \\
& \text{Projection for } \b{x}: && \b{y} = \b{\Sigma}^{-1} \b{V}^\top \b{X}^\top \b{x}, \\
& \text{Reconstruction for } \b{X}: && \widehat{\b{X}} = \b{X} \b{V} \b{V}^\top, \\
& \text{Reconstruction for } \b{x}: && \widehat{\b{x}} = \b{X} \b{V} \b{\Sigma}^{-2} \b{V}^\top \b{X}^\top \b{x}.
\end{align}

\subsubsection{Kernel Principal Component Analysis}

PCA tries to find the linear subspace for representing the pattern of data. However, kernel PCA \cite{scholkopf1997kernel} finds the non-linear subspace of data which is useful if the data pattern is not linear. The kernel PCA uses kernel method which maps data to a higher dimensional space $\b{x} \mapsto \phi(\b{x})$ where $\phi: \mathbb{R}^d \rightarrow \mathbb{R}^{d'}$ and $d' > d$ \cite{shawe2004kernel,hofmann2008kernel}. 
Note that having high dimensions has both its blessings and curses \cite{donoho2000high}. The ``curse of dimensionality'' refers to the fact that by going to higher dimensions, the number of samples required for learning a function grows exponentially. On the other hand, the ``blessing of dimensionality'' states that in higher dimensions, the representation or discrimination of data is easier. Kernel PCA relies on the blessing of dimensionality by using kernels.

The kernel matrix is $K(\b{x}_1, \b{x}_2) = \phi(\b{x}_1)^\top \phi(\b{x}_2)$ which replaces $\b{x}_1^\top \b{x}_2$ using the kernel trick.
The most popular kernels are linear kernel $\b{x}_1^\top \b{x}_2$, polynomial kernel $(1 + \langle \b{x}_1, \b{x}_2 \rangle)^i$, and Gaussian kernel $\exp(-||\b{x}_1, \b{x}_2||_2^2 / 2 \sigma^2)$, where $i$ is a positive integer and denotes the polynomial grade of kernel. Note that Gaussian kernel is also referred to as Radial Basis Function (RBF) kernel. 

After the kernel trick, PCA is applied on the data. Therefore, in kernel PCA, SVD is applied on the kernel matrix $K(\b{x}_1, \b{x}_2) = \b{U} \b{\Sigma} \b{V}^\top$ rather than on $\b{X}$. The projections of training and out-of-sample data are formulated as:
\begin{align}
& \text{Projection for } \b{X}: && \b{Y} = \b{\Sigma} \b{V}^\top, \\
& \text{Projection for } \b{x}: && \b{y} = \b{\Sigma}^{-1} \b{V}^\top K(\b{X}, \b{x}).
\end{align}
Note that reconstruction cannot be done in kernel PCA \cite{ghodsi2006dimensionality}.
It is noteworthy that kernel PCA usually does not perform satisfactorily in practice \cite{ghodsi2006dimensionality,scholkopf1998kernel} and the reason of it is the unknown perfect choice of kernels. However, it provides technical support for other methods which are explained later.

\subsubsection{Multidimensional Scaling}

Multidimensional Scaling (MDS) \cite{cox2008multidimensional} is a method for dimensionality reduction and feature extraction. It includes two main approaches, i.e., metric (classic) and non-metric. We cover the classic approach here. The metric MDS is also called Principal Coordinate Analysis (PCoA) \cite{gower1966some} in the literature. The goal of MDS is to preserve the pairwise Euclidean distance or the similarity (inner product) of data samples in the feature space. 
The solution to this goal \cite{ghodsi2006dimensionality} is:
\begin{equation} \label{equation_MDS_projection}
\b{Y} = \widehat{\b{\Lambda}}^{\frac{1}{2}} \b{V}^\top,
\end{equation}
where $\widehat{\b{\Lambda}} \in \mathbb{R}^{p \times p}$ is a diagonal matrix having the top $p$ eigenvalues of $\b{X}^\top \b{X}$ and $\b{V} \in \mathbb{R}^{n \times p}$ contains the eigenvectors of $\b{X}^\top \b{X}$ corresponding to the top $p$ eigenvalues. Comparing equations (\ref{equation_dualPCA_projection}) and (\ref{equation_MDS_projection}) shows that if the distance metric in MDS is Euclidean distance, the solutions of metric MDS and dual PCA are identical.
Note that what MDS does is converting distance matrix of samples, denoted by $\b{D}^{(\b{X})}$, to a kernel matrix $\b{K}$ \cite{ghodsi2006dimensionality} formulated as:
\begin{equation}\label{equation_MDS_kernel}
\b{K} = - \frac{1}{2} \b{H} \b{D}^{(\b{X})} \b{H},
\end{equation}
where $\b{H} := \b{I} - \frac{1}{n} \b{1} \b{1}^\top \in \mathbb{R}^{n \times n}$ is the centering matrix used for double-centering the distance matrix. When $\b{D}^{(\b{X})}$ is based on Euclidean distance, the kernel $\b{K}$ is positive semi-definite and is equal to $\b{K} = \b{X}^\top \b{X}$ \cite{cox2008multidimensional}. Therefore, $\widehat{\b{\Lambda}}$ and $\b{V}$ are the eigenvalues and eigenvectors of the kernel matrix, respectively.

\subsubsection{Isomap}

Linear methods, such as PCA and MDA (with Euclidean distance), have the lack of not capturing the possible non-linear essence of pattern. 
For example, suppose the data exist on a non-linear manifold. When applying PCA, the samples on the different sides of manifold mistakenly fall next to each other because PCA cannot capture the structure of the non-linear manifold \cite{web_Ali_Ghodsi_youtube}. Therefore, other methods are required which consider the distances of samples on the manifold. Isomap \cite{tenenbaum2000global} is a method which considers the geodesic distance of data samples on the manifold. For approximating the geodesic distances, it firstly constructs a $k$-nearest neighbor graph on the $n$ data samples. Then it computes the shortest path distances between all pairs of samples resulting in the geodesic distance matrix $\b{D}^{(\b{G})}$. Different algorithms can be used for finding the shortest paths, such as the Dijkstra and Floyd-Warshall algorithms \cite{cormen2009introduction}. Finally, it runs the metric MDS using the kernel based on $\b{D}^{(\b{G})}$ as:
\begin{equation} \label{equation_Isomap_kernel}
\b{K} = - \frac{1}{2} \b{H} \b{D}^{(\b{G})} \b{H}.
\end{equation}
The embedded data points are obtained from equation (\ref{equation_MDS_projection}) but with $\widehat{\b{\Lambda}}$ and $\b{V}$ as the eigenvalues and eigenvectors of equation (\ref{equation_Isomap_kernel}), respectively.

\subsubsection{Locally Linear Embedding}

Another perspective toward capturing the non-linear manifold of data pattern is to consider the manifold as integration of small linear patches. This intuition is similar to piece-wise linear (spline) regression \cite{marsh2001spline}. In other words, unfolding the manifold can be approximated by locally capturing the piece-wise linear patches and putting them together. For this goal, Locally Linear Embedding (LLE) \cite{roweis2000nonlinear,saul2003think} is proposed which first constructs a $k$-nearest neighbor graph similar to Isomap. Then it tries to locally represent every data sample $\b{x}_i$ using a weighted summation of its $k$-nearest neighbors. Taking $\b{w}_i$ as the $i$-th row entry of the $n \times k$ weight matrix $\b{W}$, the solution to this goal is: 
\begin{align}
& \b{w}_i = \frac{1}{\b{1}^\top \b{G}_i^{-1} \b{1}} \b{G}_i^{-1} \b{1}, \label{equation_LLE_weights} \\
& \b{G} := (\b{x}_i \b{1}^\top - \b{V}_i)^\top (\b{x}_i \b{1}^\top - \b{V}_i),
\end{align}
where $\b{G}$ is named Gram matrix and $\b{V}$ is a $d \times k$ matrix defined as $\b{V} := [\b{x}_{v_{i(1)}}, \dots, \b{x}_{v_{i(k)}}]$. The $v_{i(j)}$ denotes the index of the $j$-th neighbor of sample $\b{x}_i$ among the $n$ samples.
Note that dividing by $\b{1}^\top \b{G}_i^{-1} \b{1}$ in equation (\ref{equation_LLE_weights}) is for normalizing weights associated with the $i$-th sample so that $\sum_{j=1}^k w_{ij} = 1$ is satisfied, where $w_{ij}$ is the entry of $\b{W}$ in the $i$-th row and the $j$-th column.

After representing the samples as a weighted summation of their neighbors, LLE tries to represent the samples in the lower dimensional space (denoted by $\b{y}$) by their neighbors with the same obtained weights. In other words, it preserves the locality of data pattern in the feature space. 
The solution to this second goal (matrix $\b{Y}$) is the $p$ smallest eigenvectors of $\b{M}$ after ignoring an eigenvector having eigenvalue equal to zero. The $n \times n$ matrix $\b{M}$ is 
\begin{align}\label{equation_LLE_M}
\b{M} := (\b{I}-\b{W}')(\b{I}-\b{W}')^\top,
\end{align}
where $\b{W}'$ is considered as a $n \times n$ weight matrix having zero entries for the non-neighbor samples.

\subsubsection{Laplacian Eigenmap}

Another non-linear perspective to dimensionality reduction is to preserve locality based on the similarity of neighbor samples. Laplacian Eigenmap (LE) \cite{belkin2003laplacian} is a method which tracks this aim. This method, first, constructs a weighted graph $\b{G}$ in which vertices are data samples and edge weights demonstrate a measure of similarity such as $w_{ij} = \exp(-||\b{x}_i - \b{x}_j||_2^2 / \gamma)$. LE tries to capture the locality. It minimizes $||\b{y}_i - \b{y}_j||_2^2$ when the samples $\b{x}_i$ and $\b{x}_j$ are close to each other, i.e., weight $w_{ij}$ is large. 
The solution to this problem \cite{ghodsi2006dimensionality} is the $p$ smallest eigenvectors of the Laplacian matrix of graph $\b{G}$ defined as $\b{L} := \b{D} - \b{W}$, where $\b{D}$ is a diagonal matrix with elements $d_{ii} = \sum_j w_{ij}$. Note that according to characteristics of Laplacian matrix, for the connected graph $\b{G}$, there exists one zero eigenvalue whose corresponding eigenvector should be ignored.

\subsubsection{Maximum Variance Unfolding}

Surprisingly, all the unsupervised methods explained so far can be seen as the kernel PCA with different kernels \cite{ham2004kernel,strange2014spectral}:
\begin{align*}
\left\{
    \begin{array}{ll}
        \text{PCA:} & \b{K} = \b{X}^\top \b{X}, \\
		\text{MDS:} & \b{K} = \frac{-1}{2} \b{H} \b{D}^{(\b{X})} \b{H}, \\
		\text{Isomap:} & \b{K} = \frac{-1}{2} \b{H} \b{D}^{(\b{G})} \b{H}, \\
		\text{LLE:} & \b{K} = \b{M}^\dagger, \\
		\text{LE:} & \b{K} = \b{L}^\dagger,
     \end{array}
\right.
\end{align*}
where $\b{M}^\dagger$ and $\b{L}^\dagger$ are the pseudo-inverses of matrices $\b{M}$ and $\b{L}$, respectively. The reason of this pseudo-inverse is that in LLE and LE, the eigenvectors having smallest, rather than the largest, eigenvalues are considered. 
Inspired by this idea, another approach toward dimensionality reduction is to apply kernel PCA but with the optimum kernel. In other words, the best kernel can be found using optimization. This is the aim of Maximum Variance Unfolding (MVU) \cite{weinberger2004learning,weinberger2006unsupervised,weinberger2006introduction}. The reason for this name is that it finds the best kernel which maximizes the variance of data in the feature space. The variance of data is equal to the trace of kernel matrix, denoted by \textbf{tr}($\b{K}$), which is equal to the summation of eigenvalues of $\b{K}$ (recall that we saw in PCA that eigenvalues show the amount of variance). Supposing that $\b{K}_{ij}$ denotes the $(i,j)$ entry of the kernel matrix, the optimization problem which MVU tackles is:
\begin{equation}
\begin{aligned}
& \underset{\b{K}}{\text{maximize}}
& & \textbf{tr}(\b{K}) \\
& \text{subject to}
& & ||\b{x}_i - \b{x}_j||_2^2 = \b{K}_{ii} + \b{K}_{ij} - 2\b{K}_{ij}, \\
& & & \sum_{i=1}^n \sum_{j=1}^n \b{K}_{ij} = 0, \\
& & & \b{K} \succeq 0,
\end{aligned}
\end{equation}
which is a semi-definite programming optimization problem \cite{vandenberghe1996semidefinite}. That is why MVU is also addressed as Semi-definite Embedding (SDE) \cite{weinberger2006unsupervised}. This optimization problem does not have a closed form solution and needs optimization toolboxes to be solved \cite{weinberger2004learning}. 

\subsubsection{Autoencoders \& Neural Networks}

Autoencoders (AEs), as neural networks, can be used for compression, feature extraction or, in general, for data representation \cite{van2009dimensionality,hinton2006reducing}. The most basic form of an AE is with an encoder and decoder having just one hidden layer. The input is fed to the encoder and output is extracted from the decoder. The output is the reconstruction of the input; therefore, the number of nodes of the encoder and decoder are the same. The hidden layer usually has fewer number of nodes than the encoder/decoder layer. The AEs with less or more number of hidden neurons are called undercomplete and overcomplete, respectively \cite{goodfellow2016deep}. AEs were first introduced in \cite{rumelhart1985learning}. 

AEs try to minimize the error between input $\{\b{x}_i\}_{i=1}^n$ and decoded output $\{\widehat{\b{x}}_i\}_{i=1}^n$, i.e., reproduce the exact input using the embedded information in the hidden layer. The hidden layer in undercomplete AE is the representation of data with reduced dimensionality and compressed form \cite{hinton2006reducing}. 
Once the network is trained, the decoder part is removed and output of the innermost hidden layer is used for feature extraction from input.
To get better compression or greater dimensionality reduction, multiple hidden layers should be used resulting in deep AE \cite{goodfellow2016deep}. Previously, training deep AE faced the problem of vanishing gradients because of large number of layers. This problem was first resolved by considering each pair of layers as a Restricted Boltzmann Machine (RBM) and training it in unsupervised manner \cite{hinton2006reducing}. This AE is also referred to as Deep Belief Network (DBN) \cite{hinton2009deep}. The obtained weights are then fine tuned by backpropagation. However, recently, vanishing gradients is resolved mostly because of using ReLu activation function \cite{nair2010rectified} and batch normalization \cite{ioffe2015batch}. Therefore, the current learning algorithm used in AEs is the backpropagation algorithm \cite{rumelhart1986learning}, where error is between $\b{x}_i$ and $\widehat{\b{x}}_i$.

\subsubsection{$t$-distributed Stochastic Neighbor Embedding}

The $t$-distributed Stochastic Neighbor Embedding ($t$-SNE) \cite{maaten2008visualizing}, which is an improvement to Stochastic Neighbor Embedding (SNE) \cite{hinton2003stochastic}, is a state-of-the-art method for data visualization and its goal is to preserve the joint distribution of data samples in the original and embedding spaces.
If $p_{ij}$ and $q_{ij}$, respectively, denote the probability that $\b{x}_i$ and $\b{x}_j$ are neighbors (similar) and $\b{y}_i$ and $\b{y}_j$ are neighbors, we have:
\begin{align}
& p_{ij} = \frac{p_{j|i} + p_{i|j}}{2n}, \label{equation_TSNE_p_distribution} \\
& p_{j|i} = \frac{\exp (- ||\b{x}_i - \b{x}_j||_2^2/2\sigma_i^2)}{\sum_{k \neq i} \exp (- ||\b{x}_i - \b{x}_k||_2^2/2\sigma_i^2)}, \\
& q_{ij} = \frac{(1 + ||\b{y}_i - \b{y}_j||_2^2)^{-1}}{\sum_{k \neq l} (1 + ||\b{y}_k - \b{y}_l||_2^2)^{-1}}, \label{equation_TSNE_q_distribution}
\end{align}
where $\sigma_i^2$ is the variance of Gaussian distribution over $\b{x}_i$ obtained by binary search \cite{hinton2003stochastic}. The $t$-SNE considers Gaussian and Student's $t$-distribution \cite{student1908probable} for original and embedding spaces, respectively. The embedded samples are obtained using gradient descent over minimizing the Keullback-Leibler divergence \cite{kullback1997information} of $p$ and $q$ distributions (equations (\ref{equation_TSNE_p_distribution}) and (\ref{equation_TSNE_q_distribution})). The heavy tails of $t$-distribution gives $t$-SNE the ability to deal with the problem of visualizing ``crowded'' high-dimensional data in a low dimensional (e.g., 2D or 3D) space \cite{maaten2008visualizing,van2009learning}.

\subsection{Supervised Feature Extraction}

\subsubsection{Fisher Linear Discriminant Analysis}

Fisher Linear Discriminant Analysis (FLDA) is also referred to as Fisher Discriminant Analysis (FDA) or Linear Discriminant Analysis (LDA) in literature. The base of this method was proposed by a genius named Ronald A. Fisher \cite{fisher1936use}.
Similar to PCA, FLDA calculates the projection of data along a direction; however, rather than maximizing the variation of data, FLDA utilizes label information to get a projection maximizing the ratio of between-class variance to within-class variance. 
The goal of FLDA is formulated as the Fisher criterion \cite{welling2005fisher,xu2006analysis}:
\begin{equation}
J(\b{u}) := \frac{\b{u}^\top \b{S}_B \b{u}}{\b{u}^\top \b{S}_W \b{u}}, 
\end{equation}
where $\b{u}$ is the projection direction, and $S_B$ and $S_W$ are between- and within-class scatters formulated as \cite{welling2005fisher}:
\begin{align}
& \b{S}_B := \sum_{c=1}^{|\mathcal{C}|} n_c (\overline{\b{x}}_c - \overline{\b{x}}) (\overline{\b{x}}_c - \overline{\b{x}})^\top, \\
& \b{S}_W := \sum_{c=1}^{|\mathcal{C}|} \sum_{\b{x}_i \in c} (\b{x}_i - \overline{\b{x}}_c) (\b{x}_i - \overline{\b{x}}_c)^\top,
\end{align}
assuming that $|\mathcal{C}|$ is the number of classes, $n_c$ is the number of training samples in class $c$, $\overline{\b{x}}_c$ is the mean of class $c$, and $\overline{\b{x}}$ is the mean of all training samples.
Maximizing the Fisher criterion $J(\b{u})$ results in a generalized eigenvalue problem $\b{S}_B \b{u} = \lambda \b{S}_W \b{u}$ \cite{ghojogh2019eigenvalue}. Therefore, the projection directions (columns of the projection matrix $\b{U}$) are the $p$ eigenvectors of $\b{S}_W^{-1} \b{S}_B$ with the largest eigenvalues. Note that as $\b{S}_B$ has rank $\leq (C-1)$, we have $p \leq C-1$ \cite{murphy2012machine}.
It is noteworthy that when $n \ll d$ (e.g., in images), the $S_W$ might become singular and not invertable. In these cases, FLDA is difficult to be directly implemented and can be applied on PCA-transformed space \cite{yang2003can}.
It is also noteworthy that an ensemble of FLDA models, named Fisher forest \cite{ghojogh2017automatic}, can be useful for classifying data with different essences and even different dimensionality.  

\begin{table*}[!t]
\renewcommand{\arraystretch}{1.3}  
\caption{Some examples of applications of different methods in feature selection and extraction.}
\label{table_applications}
\centering
\scalebox{0.86}{    
\begin{tabular}{l | l || c : c}
\hline
\hline
& & \textbf{Method} & \textbf{Applications} \\
\hline
\multirow{10}{*}{\rotatebox[origin=c]{90}{\textbf{Feature Selection}}}  
& \multirow{8}{*}{Filters} 
& CC & network intrusion detection \cite{eid2013linear} \\ 
& & MI & advertisement \cite{ciesielczyk2017using}, action recognition \cite{fish2012feature} \\ 
& & $\chi^2$ Statistics & medical imaging \cite{abraham2007medical}, text classification \cite{moh2007chi,basu2012effective} \\ 
& & MB & Gaussian mixture clustering \cite{zeng2009new} \\ 
& & CBF & IP traffic \cite{williams2006preliminary}, credit scoring \cite{liu2005data}, fault detection \cite{rodriguez2007attribute}, antidepressant medication \cite{huang2009dimensionality} \\
& & FCBF & software defect prediction \cite{liu2014fecar}, internet traffic \cite{moore2005internet}, intelligent tutoring \cite{baker2007modeling}, network fault diagnosis \cite{kandula2009detailed} \\
& & Interact & network intrusion detection \cite{koc2012network}, automatic recommendation \cite{wang2013feature}, dry eye detection \cite{remeseiro2014methodology} \\
& & mRMR & health monitoring \cite{jin2012health}, churn prediction \cite{idris2013intelligent}, gene expression \cite{radovic2017minimum} \\
\cline{2-4}
& \multirow{2}{*}{Wrappers} 
& SS & satellite images \cite{jain1997feature}, medical imaging \cite{ruckstiess2011sequential} \\
& & Metaheuristic & cancer classification \cite{alba2007gene}, hospital demands \cite{jiang2017modified} \\ 
\hline
\multirow{14}{*}{\rotatebox[origin=c]{90}{\textbf{Feature Extraction}}}
& \multirow{9}{*}{Unsupervised} 
& PCA & face recognition \cite{turk1991face}, action recognition \cite{ahmad2006hmm}, EEG \cite{subasi2010eeg}, object orientation detection \cite{mohammadzade2017critical} \\ 
& & Kernel PCA & face recognition \cite{kim2002face,yang2002kernel}, fault detection \cite{choi2005fault} \\ 
& & MDS & marketing \cite{cooper1983review}, psychology \cite{robinson1995typology} \\ 
& & Isomap & video processing \cite{pless2003image}, face recognition \cite{yang2002face}, wireless network \cite{wang2009wireless} \\ 
& & LLE & gesture recognition \cite{ge2008hand}, speech recognition \cite{jain2004exploratory}, hyperspectral images \cite{kim2003hyperspectral} \\
& & LE & face recognition \cite{luo2011face,he2005face}, hyperspectral images \cite{hou2013novel} \\
& & MVU & process monitoring \cite{shao2009nonlinear}, transfer learning \cite{pan2008transfer} \\
& & AE & speech processing \cite{deng2010binary}, document processing \cite{li2015hierarchical}, gene expression \cite{chicco2014deep} \\
& & $t$-SNE & cytometry \cite{chester2015algorithmic}, camera relocalization \cite{kendall2015posenet}, breast cancer \cite{araujo2017classification}, reinforcement learning \cite{zahavy2016graying} \\
\cline{2-4}
& \multirow{5}{*}{Supervised} 
& FLDA & face recognition \cite{belhumeur1997eigenfaces,mohammadzade2018pixel}, gender recognition \cite{ghojogh2018fusion}, action recognition \cite{ghojogh2017fisherposes,mokari2018recognizing}, \\
& & & EEG \cite{subasi2010eeg,malekmohammadi2019efficient}, EMG \cite{phinyomark2012application}, prototype selection \cite{ghojogh2018principal} \\ 
& & Kernel FLDA & face recognition \cite{yang2002kernel,yang2006face}, palmprint recognition \cite{wang2006kernel}, prototype selection \cite{ghojogh2019principal} \\ 
& & SPCA & speech recognition \cite{fewzee2012dimensionality}, meteorology \cite{sarhadi2017advances}, gesture recognition \cite{samadani2013discriminative}, prototype selection \cite{ghojogh2019instance} \\ 
& & ML & face identification \cite{guillaumin2009you}, action recognition \cite{tran2008human}, person re-identification \cite{yi2014deep} \\
\hline
\hline
\end{tabular}
}
\end{table*}

\subsubsection{Kernel Fisher Linear Discriminant Analysis}

FLDA tries to find a linear discriminant but a linear discriminant may not be enough to separate different classes in some cases. Similar to kernel PCA, the kernel trick is applied and inner products are replaced by kernel function $\b{K}(\b{x}_1, \b{x}_2) = \phi(\b{x}_1)^\top \phi(\b{x}_2)$ \cite{mika1999fisher}. In kernel FLDA, the objective function to be maximized \cite{mika1999fisher} is:
\begin{equation}
J(\b{u}) := \frac{\b{u}^\top \b{M} \b{u}}{\b{u}^\top \b{N} \b{u}},
\end{equation}
where: 
\begin{align}
& \b{M} := \sum_{c=1}^{|\mathcal{C}|} n_c (\b{m}_c - \b{m}_*)(\b{m}_c - \b{m}_*)^\top \in \mathbb{R}^{n \times n}, \\
& i\text{-th }\text{entry} \text{ of } \b{m}_c (\in \mathbb{R}^{n}) := \frac{1}{n} \sum_{j=1}^{n_c} \b{K}(\b{x}_i, \b{x}_{j,c}), \\
& i\text{-th }\text{entry} \text{ of } \b{m}_* (\in \mathbb{R}^{n}) := \frac{1}{n} \sum_{j=1}^n \b{K}(\b{x}_i, \b{x}_j), \\
& \b{N} := \sum_{c=1}^{|\mathcal{C}|} \b{K}_c (I - \frac{1}{n_c} \b{1}\b{1}^\top) \b{K}_c^\top \in \mathbb{R}^{n \times n}, \\
& \text{entry } (i,j) \text{ of } \b{K}_c (\in \mathbb{R}^{n \times n_c}) := \b{K}(\b{x}_i, \b{x}_{j,c}),
\end{align}
where $\b{x}_{j,c}$ denotes the $j$-th sample in class $c$.
Similar to the approach of FLDA, the projection directions (columns of $\b{U}$) are the $p$ $(\leq C-1)$ eigenvectors of $\b{N}^{-1} \b{M}$ with the largest eigenvalues \cite{mika1999fisher}. 
The projection of data $\b{X}_t$ is obtained by $\b{U}^\top \b{K}(\b{X}, \b{X}_t)$.

\begin{table}[!t]
\renewcommand{\arraystretch}{1.3}  
\caption{Performance of feature selection and extraction methods on MNIST dataset.}
\label{table_comparison}
\centering
\scalebox{0.87}{    
\begin{tabular}{l | l | c | c | c}
\hline
\hline
& & \textbf{Method} & \textbf{\# features} & \textbf{Accuracy} \\
\hline 
\multirow{7}{*}{\rotatebox[origin=c]{90}{\textbf{Feature Selection}}}  
& \multirow{4}{*}{Filters} 
& CC & 290 & 50.90\% \\       
& & MI & 400 & 68.44\% \\ 
& & $\chi^2$ Statistics & 400 & 67.46\% \\ 
& & FCBF & 15 & 31.10\% \\    
\cline{2-5}
& \multirow{3}{*}{Wrappers} 
& SFS & 400 & 86.67\% \\
& & PSO & 403 & 59.42\% \\ 
& & GA & 396 & 61.80\% \\ 
\hline
\hline
\multirow{12}{*}{\rotatebox[origin=c]{90}{\textbf{Feature Extraction}}}
& \multirow{8}{*}{Unsupervised} 
& PCA & 5 & 60.80\% \\ 
& & Kernel PCA & 5 & 9.2\% \\ 
& & MDS & 5 & 61.66\% \\ 
& & Isomap & 5 & 75.30\% \\ 
& & LLE & 5 & 65.56\% \\
& & LE & 5 & 77.04\% \\
& & AE & 5 & 83.20\% \\
& & $t$-SNE & 3 & 89.62\% \\
\cline{2-5}
& \multirow{4}{*}{Supervised} 
& FLDA & 5 & 76.04\% \\ 
& & Kernel FLDA & 5 & 21.34\% \\ 
& & SPCA & 5 & 55.68\% \\ 
& & ML & 5 & 56.98\% \\
\hline
\hline
-- & -- & Original data & 784 & 53.50\% \\ 
\hline
\hline
\end{tabular}
}
\end{table}

\begin{figure*}[!t]
\centering
\begin{subfigure}[b]{0.3\textwidth}
\centering
\includegraphics[width=2in]{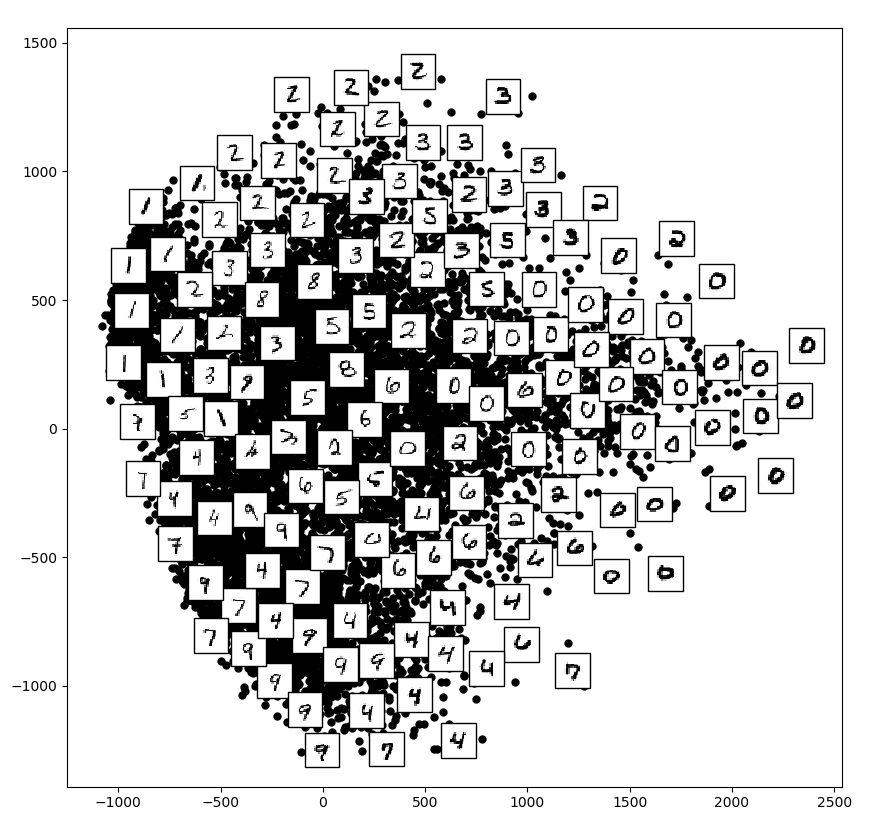} 
\caption{PCA}
\label{1}
\end{subfigure}
\begin{subfigure}[b]{0.3\textwidth}
\centering
\includegraphics[width=2in]{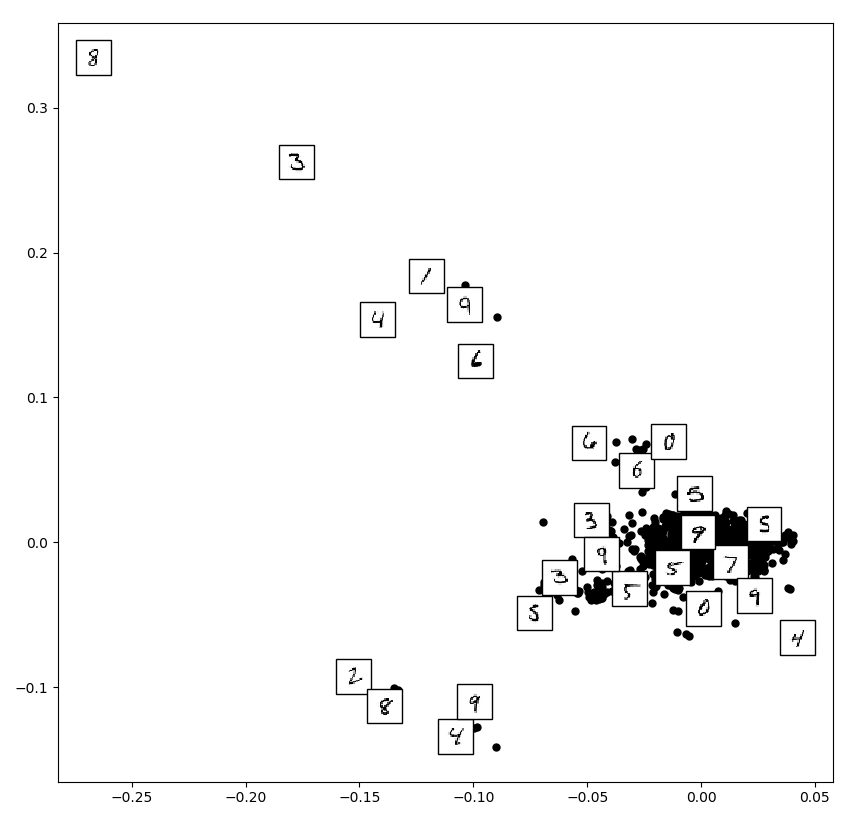} 
\caption{Kernel PCA (RBF Kernel)}
\label{1}
\end{subfigure}
\begin{subfigure}[b]{0.3\textwidth}
\centering
\includegraphics[width=2in]{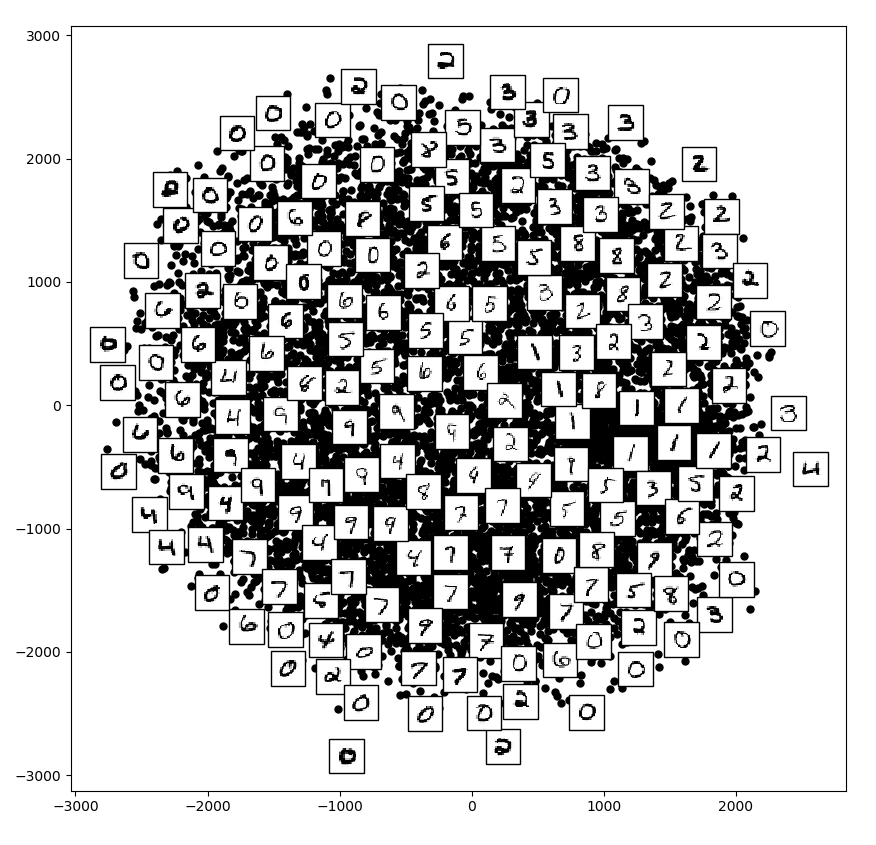} 
\caption{MDS}
\label{1}
\end{subfigure}
\begin{subfigure}[b]{0.3\textwidth}
\centering
\includegraphics[width=2in]{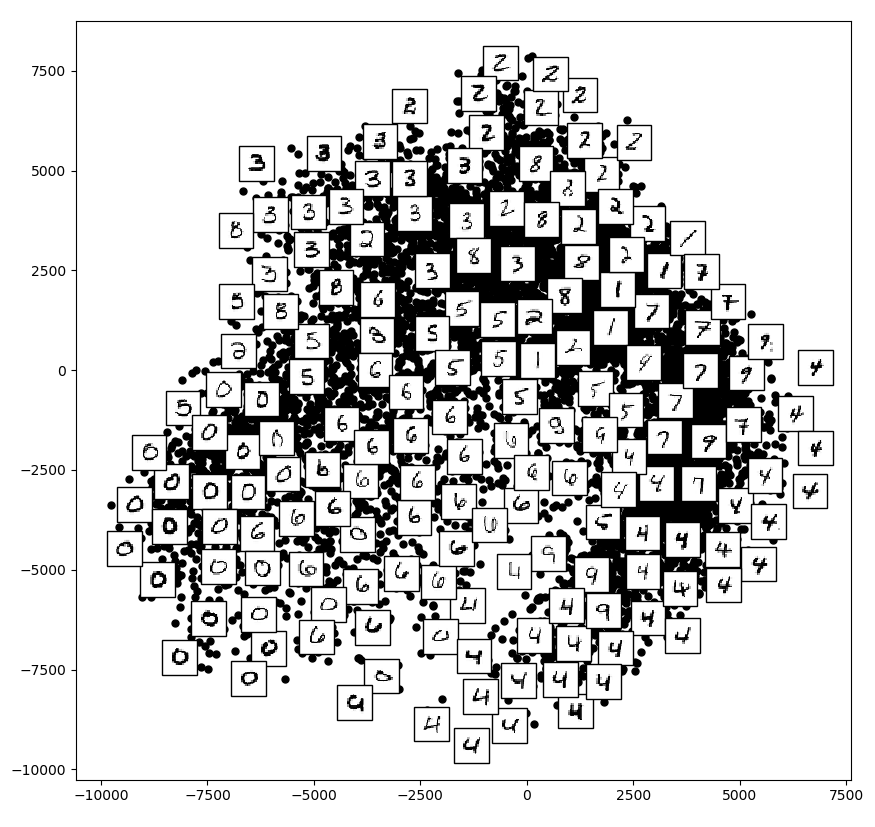} 
\caption{Isomap}
\label{1}
\end{subfigure}
\begin{subfigure}[b]{0.3\textwidth}
\centering
\includegraphics[width=2in]{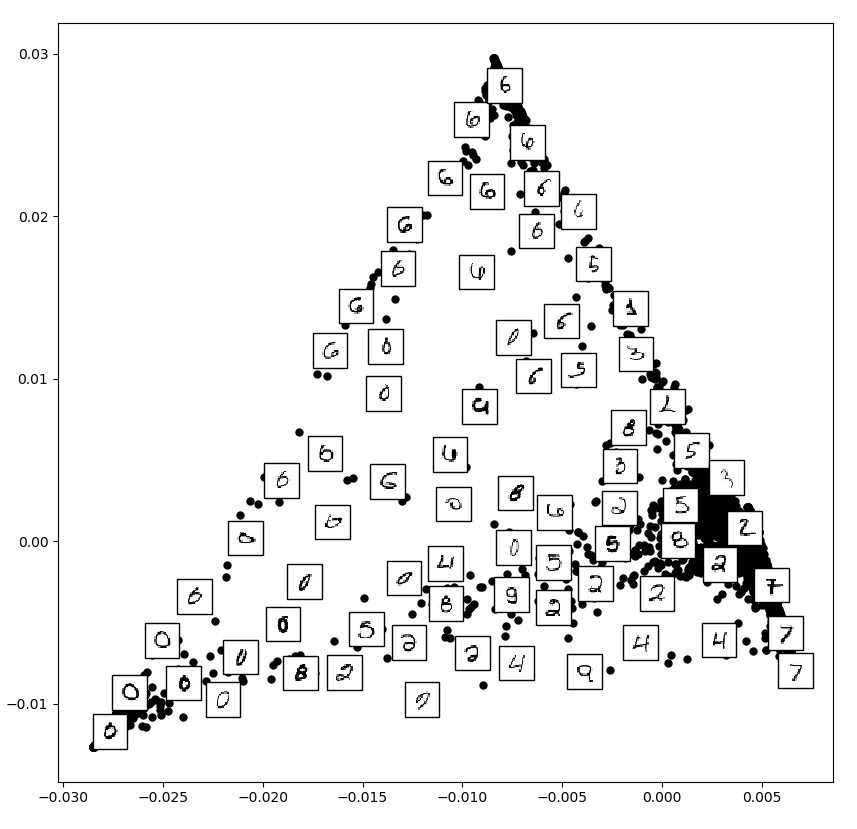} 
\caption{LLE}
\label{1}
\end{subfigure}
\begin{subfigure}[b]{0.3\textwidth}
\centering
\includegraphics[width=2in]{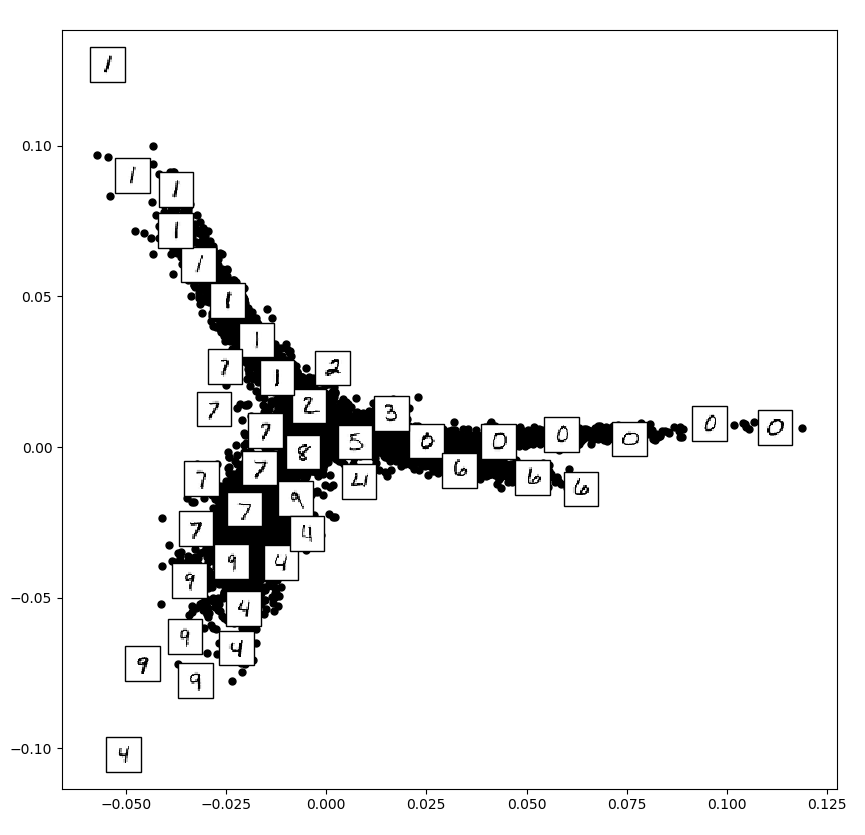} 
\caption{LE}
\label{1}
\end{subfigure}
\begin{subfigure}[b]{0.3\textwidth}
\centering
\includegraphics[width=2in]{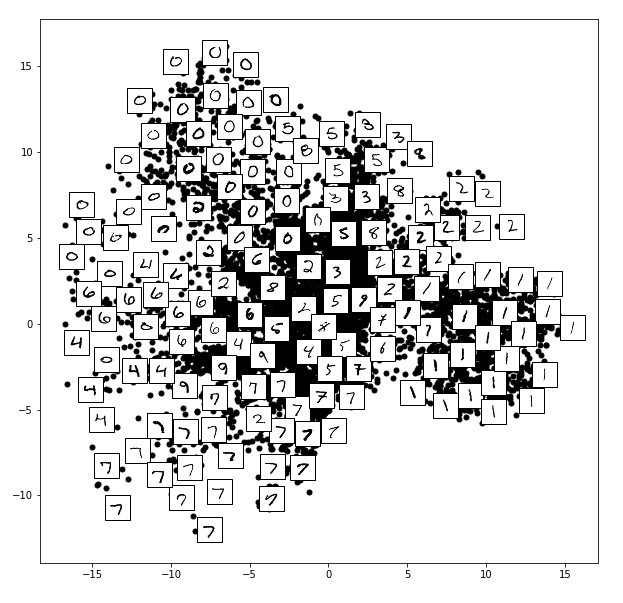} 
\caption{AE}
\label{1}
\end{subfigure}
\begin{subfigure}[b]{0.3\textwidth}
\centering
\includegraphics[width=2in]{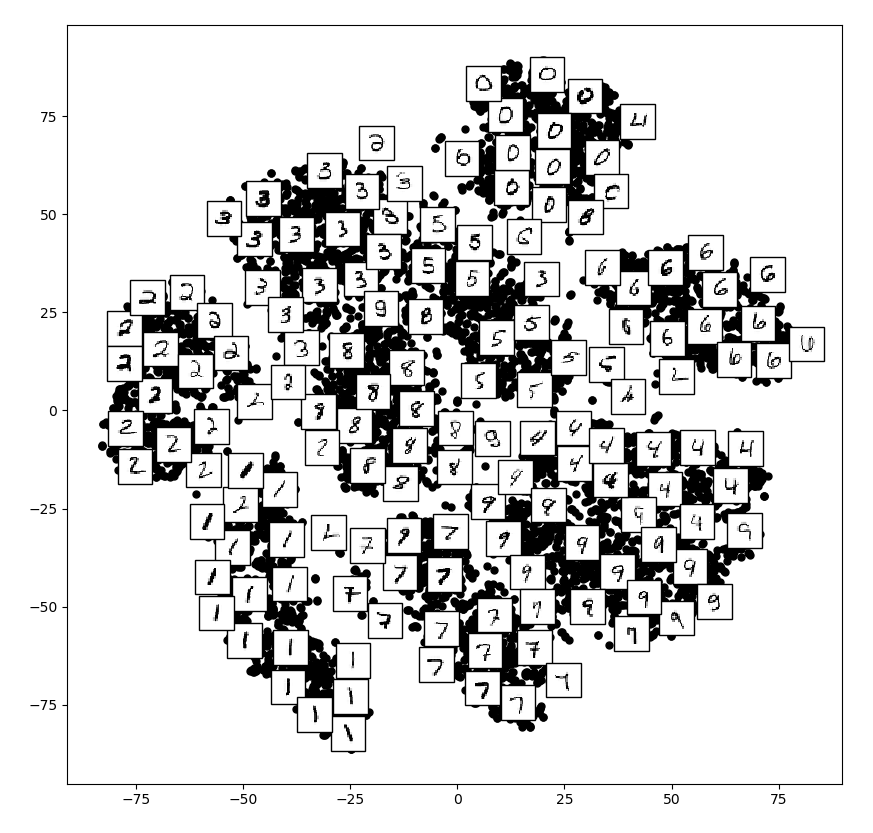} 
\caption{$t$-SNE (After PCA)}
\label{1}
\end{subfigure}
\begin{subfigure}[b]{0.3\textwidth}
\centering
\includegraphics[width=2in]{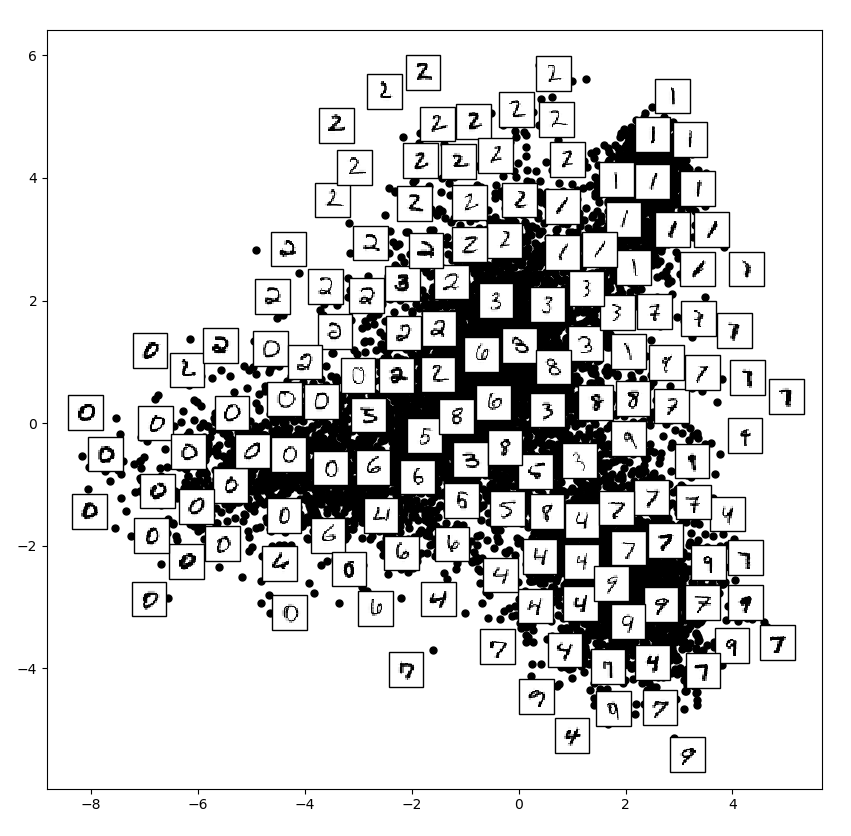} 
\caption{FLDA}
\label{1}
\end{subfigure}
\begin{subfigure}[b]{0.3\textwidth}
\centering
\includegraphics[width=2in]{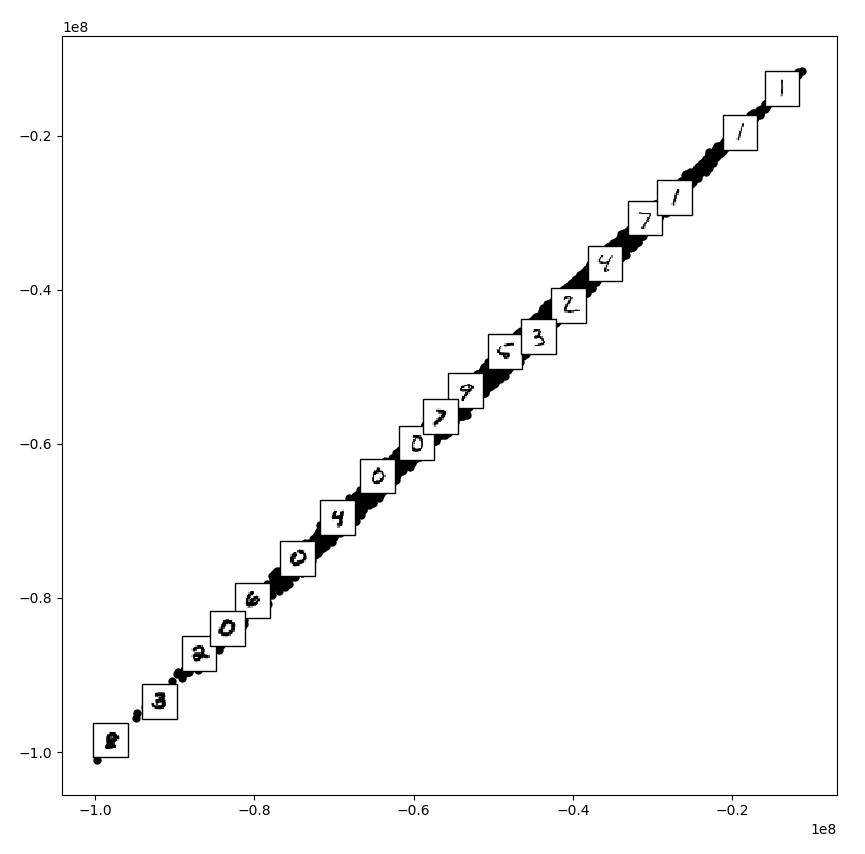} 
\caption{Kernel FLDA (Linear Kernel)}
\label{1}
\end{subfigure}
\begin{subfigure}[b]{0.3\textwidth}
\centering
\includegraphics[width=2in]{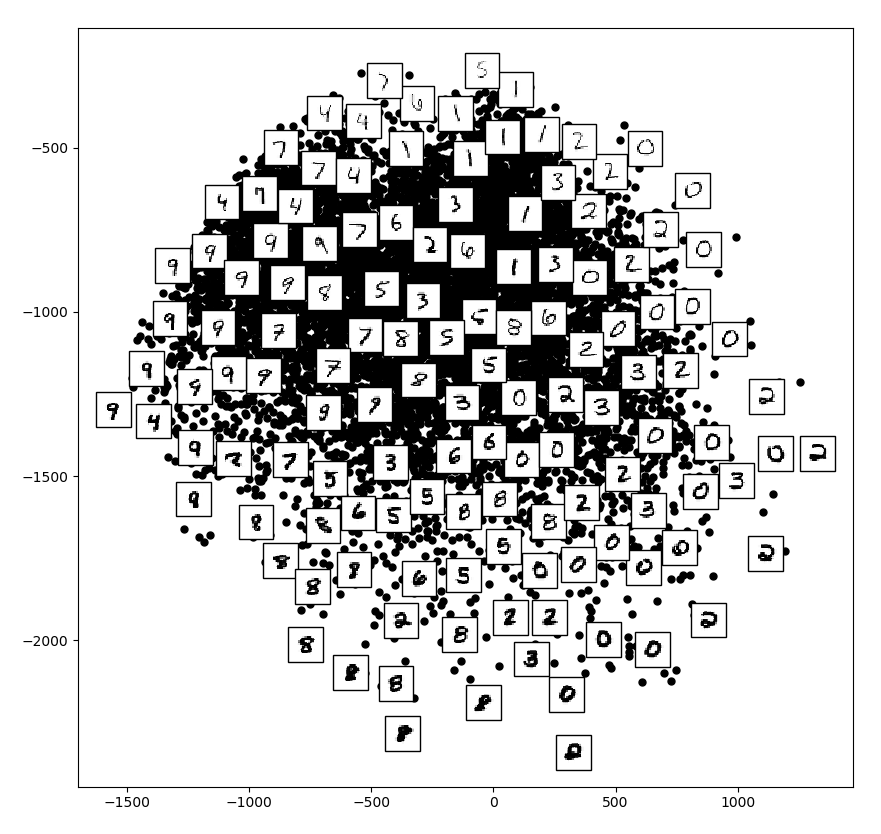} 
\caption{SPCA}
\label{1}
\end{subfigure}
\begin{subfigure}[b]{0.3\textwidth}
\centering
\includegraphics[width=2in]{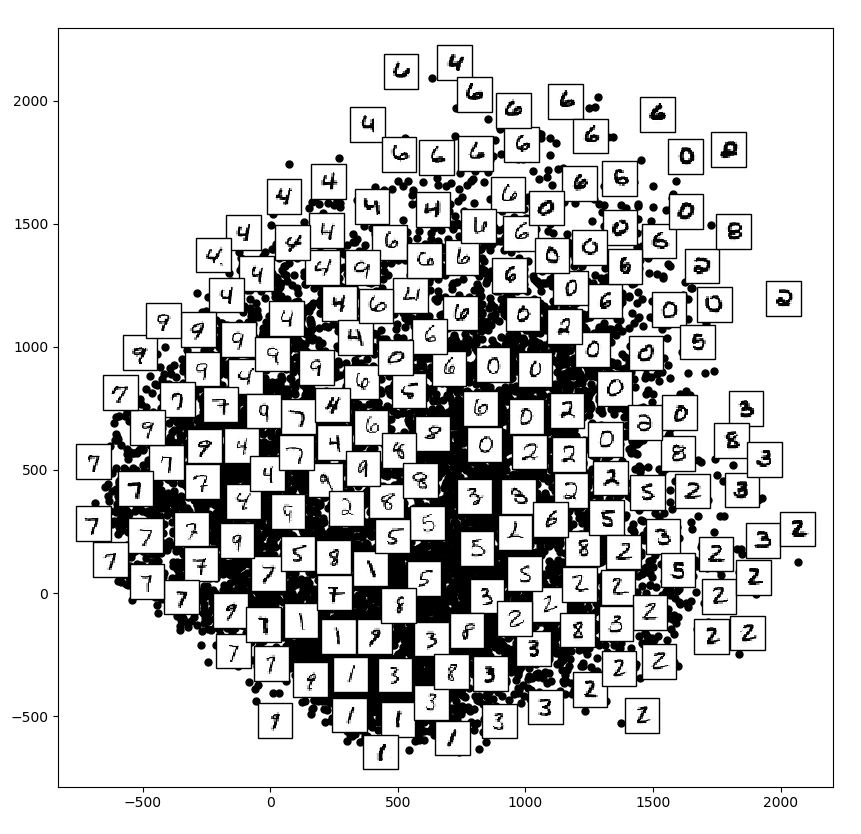} 
\caption{ML}
\label{1}
\end{subfigure}
\caption{The MNIST dataset in embedded space obtained by different feature extraction methods.}
\label{figure_embedded_space}
\end{figure*}

\subsubsection{Supervised Principal Component Analysis}

There are various approaches that have been used to do Supervised Principal Component Analysis, such as the original method of Bair's Supervised Principal Components (SPC) \cite{bair2004semi,bair2006prediction}, Hilbert-Schmidt Component Analysis (HSCA) \cite{daniuvsis2016hilbert}, and Supervised Principal Component Analysis (SPCA) \cite{barshan2011supervised}.
Here, SPCA \cite{barshan2011supervised} is presented. 

The Hilbert-Schmidt Independence Criterion (HSIC) \cite{gretton2005measuring} is a measure of dependence between two random variables. The SPCA uses this criterion for maximizing the dependence between the transformed data $\b{U}^\top \b{X}$ and the targets $\b{T} := [\b{t}_1, \dots, \b{t}_n]$. Assuming that $\b{K} = \b{X}^\top \b{U} \b{U}^\top \b{X}$ is the linear kernel over $\b{U}^\top \b{X}$ and $\b{B}$ is an arbitrary kernel over $\b{T}$, we have:
\begin{equation}
\text{HSIC} := \frac{1}{(n-1)^2} \textbf{tr}(\b{KHBH}),
\end{equation}
where $\b{H} = \b{I} - \frac{1}{n} \b{1} \b{1}^\top$ is the centering matrix.
Maximizing this HSIC criterion \cite{barshan2011supervised} results in the solution of $\b{U}$ which contains the eigenvectors, having the top $p$ eigenvalues, of $\b{XHBH}\b{X}^\top$. The encoding of data $\b{X}$ is obtained by $\b{U}^\top \b{X}$ and its reconstruction can be done by $\widehat{\b{X}} = \b{U} \b{Y}$. 

\subsubsection{Metric Learning}

The performance of many machine learning algorithms depend critically on the existence of a good distance measure (metric) over an input space, including both supervised and unsupervised learning \cite{xing2003distance}.
Metric Learning (ML) is the task of learning the best distance function (distance metric) directly from the training data. 
It is not just one algorithm but a class of algorithms. The general form of metric \cite{peltonen2004improved} is usually defined as a form similar to Mahalanobis distance:
\begin{equation}
||\b{x}_i - \b{x}_j||_{\b{A}} := (\b{x}_i - \b{x}_j)^\top \b{A}\, (\b{x}_i - \b{x}_j),
\end{equation}
where $\b{A} = \b{U}\b{U}^\top \succeq 0$ to have a valid distance metric. 
Most of the Metric Learning algorithms are optimization problems where $\b{A}$ is unknown to make data points in same class (similar pairs) closer to each other, and points in different classes far apart from each other \cite{web_Ali_Ghodsi_youtube}. It is easily observed that $||\b{x}_i - \b{x}_j||_{\b{A}} = (\b{U}^\top \b{x}_i - \b{U}^\top \b{x}_j)^\top (\b{U}^\top \b{x}_i - \b{U}^\top \b{x}_j)$, so this metric is equivalent to projection of data with projection matrix $\b{U}$ and then using Euclidean distance in the embedded space \cite{peltonen2004improved}. Therefore, Metric learning can be considered as a feature extraction method \cite{alipanahi2008distance,globerson2006metric}. The first work in ML was \cite{xing2003distance}. Another popular ML algorithm is Maximally Collapsing Metric Learning (MCML) \cite{globerson2006metric} which deals with probability of similarity of points. Here, metric learning with class-equivalence side information \cite{ghodsi2007improving} is explained which has a closed-form solution. Assuming $\mathcal{S}$ and $\mathcal{D}$, respectively, denote the sets of similar and dissimilar samples (regarding the targets), $\b{M}_\mathcal{S}$ is defined as:
\begin{equation}
\b{M}_\mathcal{S} := \frac{1}{|\mathcal{S}|} \sum_{(\b{x}_i, \b{x}_j) \in \mathcal{S}} (\b{x}_i - \b{x}_j) (\b{x}_i - \b{x}_j)^\top,
\end{equation}
and $\b{M}_\mathcal{D}$ is defined similarly for points in set $\mathcal{D}$. By minimizing the distance of similar points and maximizing the distance of dissimilar ones, it can be shown \cite{ghodsi2007improving} that the best $\b{U}$ in matrix $\b{A}$ is the $p$ eigenvectors of $(\b{M}_\mathcal{S} - \b{M}_\mathcal{D})$ having the smallest eigenvalues.

%

\section{Applications of Feature Selection and Extraction}

There exist various applications in the literature which have used the feature selection and extraction methods. Table \ref{table_applications} summarizes some examples of these applications for the different methods introduced in this paper.
As can be seen in this table, the feature selection and extraction methods are useful for different applications such as face recognition, action recognition, gesture recognition, speech recognition, medical imaging, biomedical engineering, marketing, wireless network, gene expression, software fault detection, internet traffic prediction, etc. The variety of applications of feature selection and extraction show their usefulness and effectiveness in different real-world problems. 

\section{Illustration and Experiments}

\subsection{Comparison of Feature Selection and Extraction Methods}

In order to compare the introduced methods, we tested them on a portion of MNIST dataset \cite{web_MNIST_dataset}, i.e., the first $10000$ and $5000$ samples of training and testing sets, respectively. This dataset includes images of handwritten digits with size $28 \times 28$. Gaussian Na{\"i}ve Bayes, as a simple classifier, is used for experiments in order to magnify the effectiveness of the feature selection and extraction methods. 
The number of features in feature extraction is set to five (except $t$-SNE which we use three features for the sake of visualization). In some of feature selection methods, the number of selected features can be determined and we set it to $400$ but some methods find out the best number of features themselves or based on a defined threshold. 
As reported in Table \ref{table_comparison}, the accuracy of original data without applying any feature selection or extraction method is $53.50\%$. Except for CC, FCBF, Kernel PCA, and Kernel FLDA, all other methods have improved the performance. The $t$-SNE (state-of-the-art for visualization), AE with layers 784-50-50-5-50-50-784 (deep learning), and SFS have very good results. Non-linear methods such as Isomap, LLE, and LE have better results than linear methods (PCA and MDS) as expected. Kernel PCA, as was mentioned before, does not perform well in practice because of the choice of kernel (we used RBF kernel for it).

\subsection{Illustration of Embedded Space}

For the sake of visualization, we applied feature extraction methods on the test set of MNIST dataset \cite{web_MNIST_dataset} (10,000 samples) and the samples are depicted in 2D embedded space in Fig. \ref{figure_embedded_space}. As can be seen, the similar digits almost fall in the same clusters and different digits are separated acceptably. 
The AE, as a deep learning method, and the $t$-SNE, as the state-of-the-art for illustration show the best embedding among other methods. Kernel PCA and kernel FLDA do not have satisfactory results because of choice of kernels which are not optimum. The other methods have acceptable performance in embedding.

The performances of PCA and MDS are not very promising for this dataset in discriminating the digits because they are linear methods but the data sub-manifold is non-linear. Empirically, we have seen that the embedding of Isomap usually has several legs as in octopus. Two of octopus legs can be seen in Fig. \ref{figure_embedded_space}, while for other datasets we might have more number of legs. The result of LLE is almost symmetric (symmetric triangle or square or etc) because in optimization of LLE which uses Eq. (\ref{equation_LLE_M}), the constraint is unit covariance \cite{roweis2000nonlinear}. Again empirically, we have seen that the embedding of LE usually includes some narrow string-like arms as also seen in Fig. \ref{figure_embedded_space}. FLDA and SPCA have performed well because they make use of the class labels. ML has also performed well enough because it learns the optimum distance metric.

\section{Conclusion}


This paper discussed the motivations, theories, and differences of feature selection and extraction as a pre-processing stage for feature reduction. Some examples of the applications of the reviewed methods were also mentioned to show their usage in literature. Finally, the methods were tested on the MNIST dataset for comparison of their performances. Moreover, the embedded samples of MNIST dataset were illustrated for better interpretation.

\Urlmuskip=0mu plus 1mu   

\bibliographystyle{IEEEtran}
\bibliography{references}


\end{document}